\documentclass[10pt,twocolumn,letterpaper]{article}

\usepackage{wacv}              % To produce the CAMERA-READY version
\usepackage[accsupp]{axessibility} % Improves PDF readability for those with disabilities.
\usepackage{graphicx}
\usepackage{amsmath}
\usepackage{amssymb}
\usepackage{booktabs}
\usepackage{wacv}
\usepackage{times}
\usepackage{epsfig}
\usepackage{graphicx}
\usepackage{amsmath}
\usepackage{amssymb}
\usepackage{multirow}
\usepackage{graphicx}
\usepackage{CJKutf8}
\usepackage{pifont}
\usepackage{color}
\usepackage{algorithm}  
\usepackage{algorithmicx}  
\usepackage{algpseudocode}
\usepackage[pagebackref,breaklinks,colorlinks]{hyperref}

\usepackage[capitalize]{cleveref}
\crefname{section}{Sec.}{Secs.}
\Crefname{section}{Section}{Sections}
\Crefname{table}{Table}{Tables}
\crefname{table}{Tab.}{Tabs.}

\def\wacvPaperID{971} % *** Enter the WACV Paper ID here
\def\confName{WACV}
\def\confYear{2024}

\begin{document}

%%%%%%%%% TITLE
\title{Robust Source-Free Domain Adaptation for Fundus Image Segmentation}

\author{ \textbf{Lingrui Li}$^{1,2}$, \textbf{Yanfeng Zhou}$^{1,2}$, \textbf{Ge Yang}$^{1,2}$ \thanks{Corresponding author.} \\
 $^1$Institute of Automation, Chinese Academy of Sciences  ~~ \\ $^2$School of Artifical Intelligence, University of Chinese Academy of Sciences~~ \\ 
 {\tt\small \{lilingrui2021, zhouyanfeng2020, ge.yang\}@ia.ac.cn}
}

\maketitle
%\thispagestyle{empty}
%%%%%%%%% ABSTRACT
\begin{abstract}
   Unsupervised Domain Adaptation (UDA) is a learning technique that transfers knowledge learned in the source domain from labelled training data to the target domain with only unlabelled data. It is of significant importance to medical image segmentation because of the usual lack of labelled training data. Although extensive efforts have been made to optimize UDA techniques to improve the accuracy of segmentation models in the target domain, few studies have addressed the robustness of these models under UDA. In this study, we propose a two-stage training strategy for robust domain adaptation. In the source training stage, we utilize adversarial sample augmentation to enhance the robustness and generalization capability of the source model. And in the target training stage, we propose a novel robust pseudo-label and pseudo-boundary (PLPB) method, which effectively utilizes unlabeled target data to generate pseudo labels and pseudo boundaries that enable model self-adaptation without requiring source data. Extensive experimental results on cross-domain fundus image segmentation confirm the effectiveness and versatility of our method. Source code of this study is openly accessible at https://github.com/LinGrayy/PLPB.  
% Extensive experimental results on cross-domain fundus image segmentation
% manifest the effectiveness and versatility of our method. 
% Specifically, our method outperforms state-of-the-art source-free method DPL on clean samples of C+O Dice by 3.22\% and source-dependent baseline method BEAL by 22.41\% of C+O Dice and 15.21 of ASD averaged on adversarial and clean samples. 
\end{abstract}

%%%%%%%%% BODY TEXT
\section{Introduction}
Unsupervised domain adaptation (UDA) transfers the knowledge embedded in labeled source domain data to the target domain, mitigating the problem of performance degradation caused by domain shift and dependence on expensive pixel-wise annotations~\cite{UDA1,ADPSEG,ADVENT}. However, existing UDA methods~\cite{UDA2,INTRA,FDA} mainly consider a single target domain, resulting in limited applicability in the real world. Indeed, the target domain may feature multiple data distributions, while domain adaptation techniques may need to handle samples from unseen domains. Therefore it is more practical to study open compound domain adaptation (OCDA)~\cite{open}, where the target has multiple homogeneous unlabeled domains as well as open domains unseen before.

For OCDA, two practical issues need to be addressed. 
First, existing UDA methods typically focus on solving the adaptation problem, but often do not take robustness into consideration~\cite{FDA,open,ADPSEG,AdaIN,BEAL}. 
While image degradations and/or adversarial attacks in the natural image domain have been studied extensively~\cite{r1,adv,adv2,degrad,degrad2}, their study in the medical image domain remains limited. Vulnerability of medical deep learning systems to image degradations and/or adversarial attacks may lead to bias in downstream tasks such as patient diagnosis and treatment~\cite{mia1,mia2,adv3d}. 
This raises safety concerns about the deployment of these systems in clinical settings. % However, existing UDA methods rarely considers adversarial robustness and .
Second, source data may not be available due to privacy or storage constraints.  Recent studies have developed various source-free domain adaptation techniques~\cite{DPL,SFDA,SFDA1,SFDA-FSM}. However, the performance of these techniques may drop substantially on open-domain datasets unseen before. Real-world applications require adaptation to both multiple target domains and unseen open domains. Together, this poses a more challenging problem, namely source-free open-compound domain adaptation (SF-OCDA)~\cite{SF-OCDA}, which is the subject of this study.

\begin{figure}[ht]
\centering
\scalebox{0.87}
{
\includegraphics[width=0.9\linewidth]{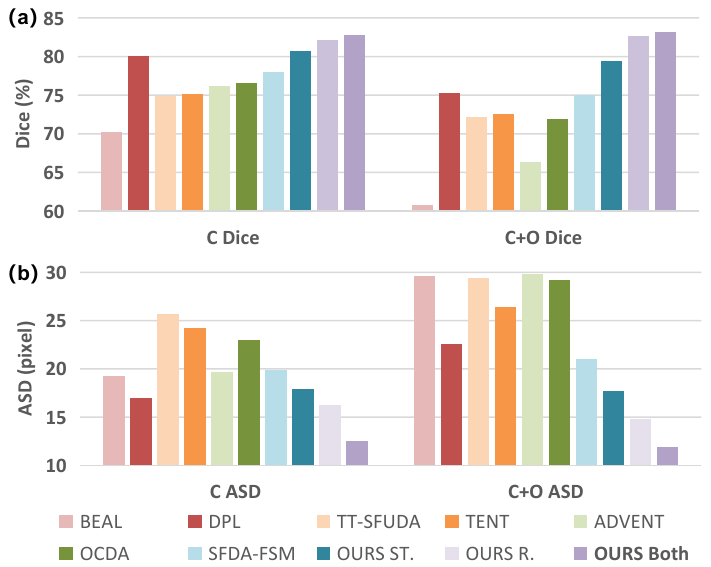}}
\caption{Performance comparison of PLPB (OURS Both) with competing methods. (a) Comparison of Dice score, higher is better. (b) Comparison of Average Surface Distance (ASD), lower is better. These results of different methods are averaged over all domain adaptation tasks for multiple clean and adversarial target datasets. Our proposed methods show substantial improvement over the baselines. ST.: only the standard source model is used in the target adaptation stage. R.: only the robust source model is used. C denotes compound target domains, C+O denotes compound and open target domains.
} \label{barchart}
\end{figure}

% In this paper, by considering the above-mentioned issues jointly, we study a novel problem: robust SF-OCDA for medical image semantic segmentation. 
In this study, we utilize adversarial sample augmentation in source training to enhance the robustness and generalization of the source model.
% , as previous works~\cite{ood,ood2} showed that a model robust to input perturbation generalizes well on out-of-distribution data. 
And in the target model training stage, we propose a \textbf{p}seudo-\textbf{l}abeling and \textbf{p}seudo-\textbf{b}oundary (PLPB) method for SF-OCDA. Our PLPB method utilizes pseudo labelling, pseudo boundary modeling and entropy minimization to enable effective self-training in the target domain without source data.
% Specifically, our method uses the information of pseudo boundaries to improve the prediction performance on soft boundary regions of the target domain images, as modeling the boundaries helps the model generalize better to unseen target domains and generate more precise and coherent predictions~\cite{BEAL}. 
By adversarial sample augmentation in the source domain and explicit modeling of the boundaries in the target domain, our model generalizes better to the unseen target domain.
In addition, our work uses adversarial samples in domain adaptation and leverages them to benchmark model performance. 
Experimental results show that our method substantially enhances the robustness while maintaining model adaptation performance on clean samples in target domains. Figure \ref{barchart} shows the overall performance of our method on clean and adversarial samples in comparison with competing methods.
% Similar to the work in~\cite{r1}, we study three different cases of model availability: given (i) only the standard source model; (ii) only the robust source model; (iii) \textit{Both} (standard+robust) models. The best strategy is training two models: a robust and a standard model in the source domain, and then adapting the robust model to the target domain but with the pseudo labels and pseudo boundaries generated by the standard model. 
% Furthermore, when the gap between source and target domain is trivial, using non-robust or robust pseudo parts does not cause big performance difference. Otherwise non-robust counterparts are suggested.
The contributions of our work are summarized as follows:
\begin{itemize}
\item{We proposed a two-stage training strategy for robust domain adaptation for semantic segmentation of medical images without source data.}

% \item{We propose a denoised pseudo-labeling and pseudo-boundary approach, PLPB, to solve SF-OCDA. And we evaluate the efficacy of PLPB on two public fundus datasets and one open domain dataset which are popular benchmarks for UDA tasks, demonstrating improvements on both clean and robust samples, especially on open dataset.}
\item{We utilize a pseudo-boundary loss in the target adaptation stage and develop a new domain adaptation method PLPB. Our method models edge information and achieves good performance on Average Surface Distance (ASD) metric and obtains precise boundary prediction. Without requiring source data, our method achieves comparable and sometimes higher performance than state-of-the-art (SOTA) source-dependent UDA methods and other SF-DA methods.}

\item{We evaluate the efficacy of PLPB on two public fundus datasets and one open domain which are popular benchmarks for UDA tasks, demonstrating improvements on both clean and adversarial samples. By utilizing adversarial samples, our method also demonstrates good generalization capability in the open domain. Our method is flexible and can be combined with other existing adaptation techniques.}
\end{itemize}

%------------------------------------------------------------------------

\section{Related Works}

%-------------------------------------------------------------------------
\subsection{Unsupervised Domain Adaptation (UDA)}
Unsupervised domain adaptation (UDA) aims to tackle domain shift by adapting the training process of a model in an unsupervised manner. Leveraging unsupervised learning reduces the laborious and time-consuming data labeling work for the target domain. Therefore, UDA is a promising method to solve domain shift problems, especially in the medical field whose data is diverse and requires expert data labeling.

Recently, many deep learning-based domain adaptation methods\cite{UDA1,UDA2,UDA3} have been proposed and achieved encouraging results. Many methods tackle the domain shift issue by extracting invariant features and a typical approach is adversarial learning\cite{ADPSEG,ADVENT}. Another popular method is image-to-image translation\cite{AdaIN,FDA}. 
Noticeably, image-to-image translation usually introduces artifacts, which may be not a proper approach in the medical field. 
Furthermore, although adversarial learning can align the latent feature distribution of different domains, the results of multiple adversarial learning-based methods are easily suffering from sub-optimal performance due to the difficulty of stabilizing the training process of multiple adversarial modules\cite{CADA}. 

Therefore, we neither use adversarial learning nor image-to-image translation in the adaptation stage. As boundary is domain-invariant information in different domains, modeling the boundaries helps the model generalize better to unseen target domains and generate more precise and coherent predictions, we utilize the low-level boundary information in both source training and target training stages. The recent work\cite{isbi2022} also considers low-level edge information by edge map. However, it does not consider the practical reality of source-free setting and open target domains.

\begin{figure*}[htbp]
\centering
\scalebox{0.85}
{
\includegraphics[width=1\textwidth]{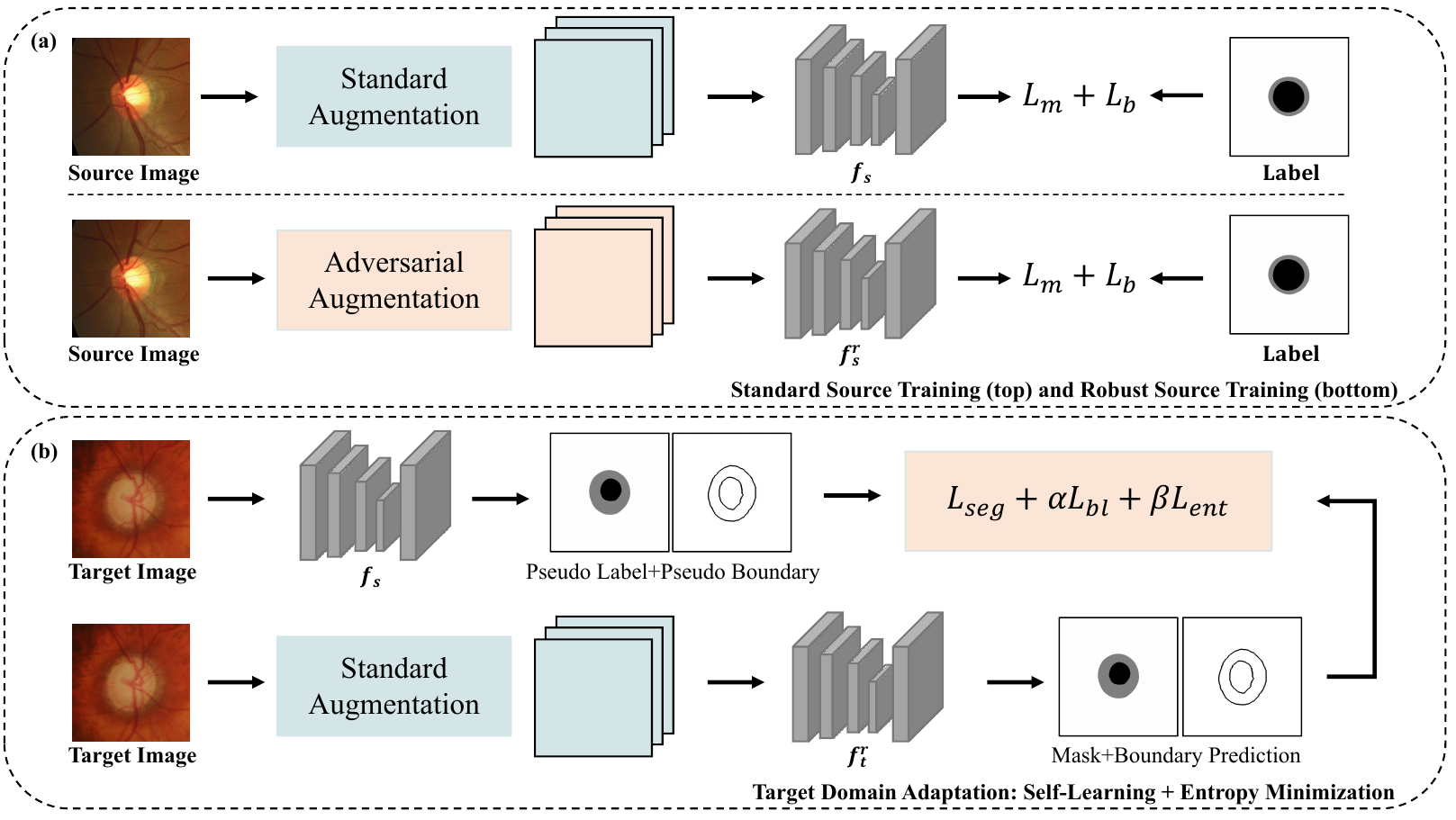}}
\caption{\textbf{Overview of our proposed method \textit{pseudo labeling and pseudo boundary} (PLPB) for source-free open compound domain adaptation.} $f_s$ denotes the standard source model. $f_s^r$ denotes the robust source model. Standard Augmentation includes several typical image augmentations which are described in section~\ref{details}, while Adversarial Augmentation means generating adversarial samples by Projected Gradient Descent (PGD)~\cite{PGD}. The training of the robust source model uses the source data, corresponding labels, and adversarial examples.
The target model only uses the target data and the source models and combines three losses with pseudo-label and pseudo-boundary. Robust target model $f_t^r$ is initialized from $f_s^r$.
} \label{fig-twostages}
\end{figure*}

%-------------------------------------------------------------------------
\subsection{Source Free Domain Adaptation (SF-DA)}
Existing SF-DA methods can be mainly divided into two categories: (i) generating images bypassing the dependence on the source data~\cite{SFDA,SFDA1,SFDA2}; (ii) self-supervision with target pseudo-labels~\cite{DPL,SHOT}. The generative approach is often difficult to scale up, as learning to generate the images is difficult. On the other hand, pseudo-label based methods are easy to handle and have recently provided very promising results. Thus, in this work, we propose to leverage the self-supervised method as in \cite{DPL,TTSFUDA}.

In SF-DA, without access to the source data, only the source pre-trained model is provided in the target training stage. SHOT \cite{SHOT} maintains the source hypothesis by fixing the trained classifier and maximizes mutual information of target outputs for distribution alignment. Lately, Liu et al.\cite{SFDA1} introduce source-free domain adaptation for semantic segmentation and utilize the batch normalization statistics of the source model to recover source-like samples. 

Though recent SF-DA methods have made great progress in object recognition and semantic segmentation, SF-DA has not been well-investigated in medical image segmentation. Moreover, they simply apply the source model as a pre-trained model to generate pseudo labels for unlabeled target data, ignoring useful information embedded in the boundaries of the objects which is important in medical images. In contrast, we utilize boundary information stored in the source model to generate pseudo boundaries, thus making better use of the source model. 

\vspace{-0.05cm}
In this work, we consider the source-free open compound domain adaptation (SF-OCDA)\cite{SF-OCDA} for robust fundus semantic segmentation, extending SF-DA to a more realistic setting which includes an open target domain.
SFDA-FSM\cite{SFDA-FSM} and DPL\cite{DPL} also adopt two-stage training strategies and investigate domain adaptation in semantic segmentation. Nevertheless, the main difference between our method and the above two methods lies in the setting. First, both SFDA-FSM\cite{SFDA-FSM} and DPL\cite{DPL} focus on the single or close-set target domain adaptation, where the performance on only one target domain is evaluated. Our SF-OCDA setting focuses on both the unlabeled compound domain and the unseen open domain, which is more challenging and practical than SFDA-FSM\cite{SFDA-FSM} and DPL\cite{DPL}. Second, both methods do not take adversarial attacks into consideration. 
% The comparison between SF-OCDA and existing adaptation settings is reported in Tab. 
%-------------------------------------------------------------------------
\subsection{Robust Training}
A wide variety of attack mechanisms \cite{PGD,adv,adv2} has been proposed since the vulnerability of deep network was shown first by \cite{goodfellow2014}. This also has led to the development of strategies to defend against such attacks, called defense mechanisms \cite{advmia,r1}. Among them, adversarial training has stood out as the most reliable way to train robust models. 
% We follow the adversarial training method as in \cite{PGD} because it is effective, fast, and easy to implement. 
%-------------------------------------------------------------------------
\subsection{Robust Domain Adaptation}
Our work is also inspired by recent studies on robust transfer learning\cite{robust,advtl2,r1}. A notable work \cite{advtl} shows that robust source feature extractors can be effective in preserving robustness, while maintaining sufficiently high accuracy on clean samples. On the other hand, \cite{r1} shows that the robust pre-trained models not only perform well on targets without adversarial training, but also improve the accuracy on clean samples. Additionally, ~\cite{ood,ood2} shows that a pre-trained model that is more robust to input perturbation provides a better initialization for generalization on downstream out-of-distribution data. These results support the hypothesis that robust models also transfer better. However, the existing methods are neither developed nor tested in the settings of SF-OCDA. Peshal et al.~\cite{r1} studied the problem of SF-DA under the adversarial perturbations in image classification. However, pixel-level segmentation is more challenging and fundamentally different from the image-level classification which just associates the label with a whole image~\cite{SFDA1}.
% Many applications of UDA also require models to be robust against data perturbations~\cite{adv,adv2}. Although adversarial samples may not occur in naturally acquired data, utilizing them present new opportunities for medical imaging researchers to investigate their models, with the ultimate goal of increasing robustness and optimizing the decision boundaries for different tasks. Robustness evaluation estimates potential failure probabilities when the model is pushed to its limits~\cite{robust} and adversarial samples are theoretically proven to be helpful for generalization on out-of-distribution data\cite{ood, ood2}. 

%------------------------------------------------------------------------
\section{Methods}

Given the source dataset ($x_s^i$, $y_s^i$) $\sim$ $D_s$ from the source domain, we first train a source model $f_s$ on $D_s$. Given $f_s$, our final goal is to obtain a target model $f_t$ that can accurately segment target images $x_t^i$ from target domain $D_t$, while being robust to adversarial examples $x_t^{adv}$ simultaneously. 

\begin{figure}[ht]
\centering
\scalebox{1}
{
\includegraphics[width=1\linewidth]{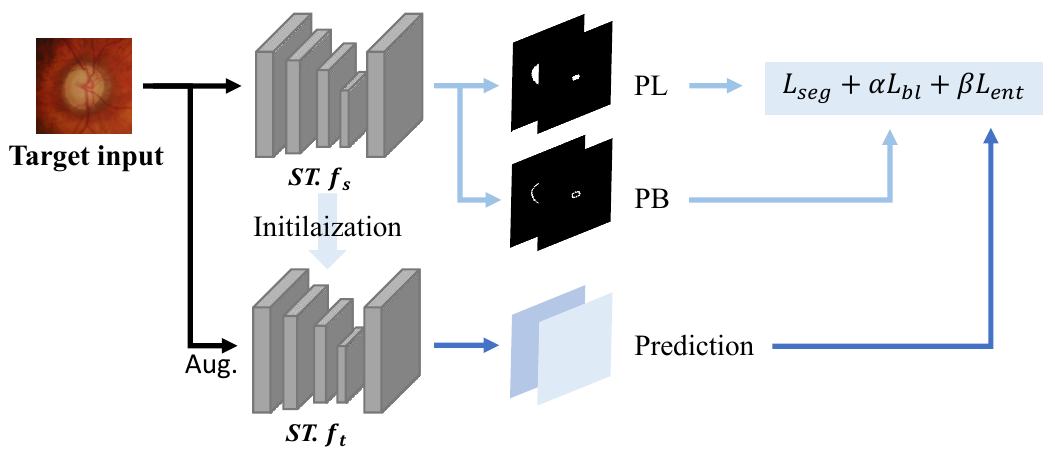}}
\caption{\textbf{Overview of standard flow on target adaptation stage.} Aug. denotes augmentation we used in target training which is described detailedly in section~\ref{details}. PB denotes pseudo-boundary. PL denotes pseudo-label. This is the case of giving only the standard source model, i.e. standard target adaptation. The only difference of robust flow is that the robust source model is given instead of the standard model.
} \label{fig1}
\end{figure}

The entire procedure consists of two steps, shown in Figure \ref{fig-twostages}. Step (i), as shown in Figure \ref{fig-twostages}(a), we train source models on the source domain in a supervised way. We train two models in the source domain, i.e., a standard model and a robust model, following standard and robust training processes, respectively, where the robust training data is augmented with adversarial examples. Step (ii), we initialize the target models with the aforementioned source models, and then adapt the source model to the target domain without source data, using pseudo-labels and pseudo-boundaries generated from the source model, as shown in Figure \ref{fig-twostages}(b).
The pseudo-labels generated from the standard and robust source models are referred to as non-robust labels and robust labels for simplicity. The same goes for the definitions of non-robust boundaries and robust boundaries. Figure ~\ref{fig1} illustrates our framework of target model training for fundus image segmentation
% , which is a multi-label semantic segmentation problem,
via denoised pseudo-labeling and pseudo-boundary (PLPB) with entropy minimization in SF-OCDA. In the following, We first describe the supervised training in the source domain in Section \ref{3.1}, then we propose the pseudo-boundary strategy and generating pseudo-labels on the target adaptation stage in Section \ref{3.2}.

\subsection{Source Domain Model Training}\label{3.1}
Similar to~\cite{BEAL}, our segmentation network includes a boundary prediction branch to regress the boundary, and a mask prediction branch to produce mask prediction for each input image.
Previous methods often perform unsatisfactorily in the accurate prediction of boundaries in the target domain, particularly in areas with indistinct boundaries between different structures, as a result of domain shift between source and target domain, combined with low-intensity contrast between the various structures. 
Incorporating boundary loss has been demonstrated to be an effective approach in mitigating the performance degradation caused by inaccurate prediction of boundaries in the target domain~\cite{BEAL}.

In the source domain, we train a deep neural network $f_s$ by minimizing the standard binary cross-entropy loss and boundary loss. Given a labeled image from the source domain, we denote $\hat{y}_m^i$ and $\hat{y}_b^i$ as its mask prediction and boundary prediction from the source model, and $y_m^i$ and $y_b^i$ as the ground truth of the mask and boundary, respectively. To optimize the network, we calculate the mask prediction loss $L_m$ and the boundary regression loss $L_b$ on the source domain images. The total loss of the segmentation network in the source domain is:
\noindent 
% \begin{equation}
% L = L_m+L_{b},\\
% % \end{equation}
% % \begin{equation}
% L_m = L_{BCE}(y^i_m,Y_m^i),
% \\
% and L_{b} = \frac{1}{N}\sum_i(y_b^i-Y_b^i)^2.
% \end{equation}
\begin{align}
\nonumber L = L_m& +L_{b},\\
\nonumber L_m = L_{BCE}&(y^i_m,\hat{y}_m^i),\\
% \nonumber L_m = -\frac{1}{N} [y^i_m \cdot log(\hat{p_v})+(1-y_{PL_v})\cdot log(1-\hat{p_v})],\\
\text{and } L_{b} = \frac{1}{N}&\sum_i(y_b^i-\hat{y}_b^i)^2.
\end{align}

Besides a standard source model $f_s$, we also train a robust source model $f^r_s$. Specifically, the only difference of $f^r_s$ is that we augment the training data with adversarial examples generated by Projected Gradient Descent (PGD)~\cite{PGD} for robust training to improve robustness.

\begin{table*}[ht]
\caption{Quantitative comparison of different methods on the target domain datasets, including both adversarial and clean images. (Note: D denotes Dice, R. stands for robust, i.e. adversarial metrics that are tested on the adversarial samples, - means the result is not reported by that method, S-F denotes source-free.) All our methods have higher adversarial values compared to the baselines. And the performance of our methods on clean samples is comparable and mostly higher than the other methods. The best value is presented in bold and the second best is underlined. (D $\uparrow$: \%, ASD $\downarrow$: pixel.)}\label{tab1}
\centering
\scalebox{0.73}
{
\begin{tabular}{l|c|c|c|c|c|c|c|c|c|c}

\hline \multirow{2}{*}{ Dataset } & \multirow{2}{*}{ Method } & \multirow{2}{*}{ S-F } & \multicolumn{4}{|c|}{ Optic disc segmentation } & \multicolumn{4}{|c}{ Optic cup segmentation } \\
\cline { 4 - 11 } & & & $\mathrm{D}[\%]$ & ASD & R.D[\%] & R.ASD & $\mathrm{D}[\%]$ & ASD & R.D[\%] & R.ASD \\

\hline
 & W/o adaptation &  &  83.02&	23.36	&-&-&	73.10&	13.87&	-&	-\\
 & Oracle &  & 95.74	&6.05	&-&	-	&83.97	&5.38&	-	&-\\
\hline
 & AdvEnt~\cite{ADVENT} & \textcolor{red}{\ding{56}} &86.76	&12.36&	71.30&	35.48&	74.48&	11.43&	46.01&	43.07 \\
 & BEAL~\cite{BEAL} & \textcolor{red}{\ding{56}} & 87.54&	19.96&	49.77&	27.04&	74.62&	12.88&	38.32&	\underline{29.95}\\
 & OCDA~\cite{open}& \textcolor{red}{\ding{56}} &86.47&16.76 	&62.86&49.17 &	76.74&10.94 &	52.11&53.76 
 \\
 & DPL~\cite{DPL}& \textcolor{green}{\ding{52}} &  90.13&	9.43&	72.78	&35.43&	76.78&	\textbf{9.01}	&52.29&	40.85 \\
\multirow{2}{*}{ RIM-ONE-r3 } & TT-SFUDA~\cite{TTSFUDA} & \textcolor{green}{\ding{52}} & 85.00&	17.05&	62.84&	61.22&	76.62&	10.31&	43.60	&61.21 \\
 & TENT~\cite{TENT} & \textcolor{green}{\ding{52}} & 82.92&	23.63&	61.72&	50.47&	72.95&	14.00&	45.70&	52.92\\
 & SFDA-FSM~\cite{SFDA-FSM} & \textcolor{green}{\ding{52}} & 	82.98&	23.69 	&73.60&	37.59 &	73.56&	14.51 	&54.27&	30.65 
\\
 & Ours (Standard Source) & \textcolor{green}{\ding{52}} &\underline{92.43}	&\underline{6.96}	&74.35&	35.58&	\underline{77.82}&	\underline{10.03}&	55.91	&45.40\\
 & Ours (Robust Source) & \textcolor{green}{\ding{52}} & 91.39&	8.07	&\textbf{77.22}&	\underline{23.63}&	75.59	&11.13&	\textbf{60.34}	&50.11\\
&\textbf{Ours (Both)} & \textcolor{green}{\ding{52}} & \textbf{92.89}&\textbf{6.52}&	\underline{76.20}&	\textbf{22.92} &	\textbf{77.94}	&10.07	&\underline{60.13}	&\textbf{26.83}\\
\hline
\multirow{14}{*}{Drishti-GS } &W/o adaptation &  &  94.04&	7.47	&-	&-&	80.22&	13.47&	-&	-\\
% Oracle~\cite{BEAL} &  & 97.40&	5.89&	-&	-&	90.10&	11.50&	-&	-\\
&Oracle &  & 97.40&	3.58& -&	-&	90.10&	9.50&	-&	-\\
\hline
&AdvEnt~\cite{ADVENT} & \textcolor{red}{\ding{56}} & 95.85	&4.65	&91.18	&10.81&	78.27&	14.71	&65.52	&25.29\\
&BEAL~\cite{BEAL} & \textcolor{red}{\ding{56}} &95.74&	8.32	&72.34&	20.57&	\textbf{86.45}&	16.65	&57.47&	18.47\\
&OCDA~\cite{open}& \textcolor{red}{\ding{56}} &95.41&5.56 &	91.63&11.43 &	79.95&13.68 &	67.56&22.98 
\\
&DPL~\cite{DPL} & \textcolor{green}{\ding{52}} &  \underline{96.39}&	4.88&	93.02&	9.58&	83.53&11.39&	75.99	&16.35 \\
&TT-SFUDA~\cite{TTSFUDA} &  \textcolor{green}{\ding{52}}& 95.22&	6.00&	91.64&	12.27	&80.67&	13.00&	64.01&	24.01 \\
&TENT~\cite{TENT} & \textcolor{green}{\ding{52}} & 94.06&	7.56&	91.32&	13.32&	80.12&	13.52	&72.17	&18.77\\
&SFDA-FSM~\cite{SFDA-FSM} & \textcolor{green}{\ding{52}} & 	93.83&	7.76 &	90.26&	17.22 &	83.19&	11.95 &	70.21&	16.09 
\\
&Ours (Standard Source)& \textcolor{green}{\ding{52}} & 96.01&\underline{4.70} &	92.05&11.25 &	\underline{83.71}&\textbf{10.91} &	73.92&18.15 
 \\
&Ours (Robust Source) & \textcolor{green}{\ding{52}} & 95.67&5.09 	&\textbf{95.55}&\textbf{5.13}&	82.87&11.55 &	\underline{78.08}&\underline{15.11}\\
&\textbf{Ours (Both)} & \textcolor{green}{\ding{52}} & \textbf{96.51}&\textbf{4.01} &	\underline{95.29}& \underline{5.25 }	&83.56&\underline{11.11} &	\textbf{80.02}&\textbf{13.69}
 \\
\hline
\multirow{14}{*}{REFUGE val (Open)} &W/o adaptation & &  70.36&52.75 &	-&	- & 71.79&26.09  &	-&	-\\
&Oracle &  &95.47&6.14 & - &	- &	88.82&	4.20 & - &	- \\
\hline
&AdvEnt~\cite{ADVENT} & \textcolor{red}{\ding{56}} &67.73&35.59 &	41.25&53.65 &	60.93&26.10 &	16.47&85.11 
 \\
&BEAL~\cite{BEAL} &\textcolor{red}{\ding{56}}  & 72.21&	52.94&	24.47&	58.67&	64.64&	47.34	&5.35&	42.79\\
&OCDA~\cite{open}& \textcolor{red}{\ding{56}} &85.38&29.70 &	48.71&58.98 &	79.14&14.20 &	37.44&63.59 
 \\
&DPL~\cite{DPL}& \textcolor{green}{\ding{52}} &  85.48&	8.23&	52.74&	54.88&	72.14&	15.11&	41.33&	59.12
 \\
&TT-SFUDA~\cite{TTSFUDA} & \textcolor{green}{\ding{52}} & 82.60&	33.00	&50.80&	51.73	&77.86&	12.56	&41.61&	56.53
 \\
&TENT~\cite{TENT} & \textcolor{green}{\ding{52}} & 67.73&	35.59&	41.25&	53.65&	60.93&	26.10	&16.47&	85.11\\
&SFDA-FSM~\cite{SFDA-FSM} & \textcolor{green}{\ding{52}} & 	80.64&	8.33 	&77.27&	35.39 &	78.19&	8.91 	&40.03&	39.80 
\\
&Ours (Standard Source) & \textcolor{green}{\ding{52}} & 91.54&6.87 &	88.51&17.68 &	79.78&7.44 &	46.46&37.46 
\\
&Ours (Robust Source) & \textcolor{green}{\ding{52}} & \underline{91.86}&\underline{6.77}&	\underline{91.11}&\underline{8.40}	&\textbf{80.40}&\underline{7.14}	&\textbf{72.31}&\underline{25.71}

\\
&\textbf{Ours (Both)} & \textcolor{green}{\ding{52}} & \textbf{92.53}&\textbf{6.54} &\textbf	{91.35}&\textbf{8.11} &	80.31&\textbf{7.12}& 	\underline{71.05}&\textbf{20.87}
\\
\hline
\end{tabular}
}
\end{table*}

\subsection{Target Domain Model Training}\label{3.2}
In the target domain, the only available components are source models and target data $x_t^i \sim D_t$. We adapt the standard source model $f_s$ and robust source model $f^r_s$ to the target domain separately so that we obtain two target models $f_t$ and $f_t^r$. %Different with~\cite{r1}, 
Standard and robust target models are trained in the same way, only with different initializations. In this section, we only describe details of standard target model training for simplicity.

\subsubsection{Segmentation Loss with Pseudo Labels} 
As we do not have the ground truth of target data, we generate the pseudo labels $y_{PL}$ and pseudo boundaries $y_{PB}$ from the aforementioned source model. 
Given an unlabeled image from the target domain, we denote $p_v$ as the mask prediction probability on $v$-th pixel obtained from the source model, then use $\hat{p}_v$ and $\hat{b}_v$ to denote the mask probability and boundary prediction obtained from the target model.
The pseudo label can be generated as: $y_{PL_v} = 1[ p_v \ge t ]$,
% \noindent 
% \begin{equation}
% y_{PL_v} = 1[ p_v \ge t ],
% \end{equation}
where $1(\cdot)$ is the indicator function, $t \in(0, 1) $ is a probability threshold to generate binary pseudo labels for the segmentation task. And we alleviate the noise of the pseudo labels by using the
% uncertainty estimation and prototype estimation to generate the
label selection mask $m_v$~\cite{DPL}, in which a pseudo label is selected when the network’s uncertainty on prediction is low and the encoded feature lies closer to the object prototype than the background prototype:
\noindent
\begin{equation}
L_{seg} = -\sum_v m_v * [y_{PL_v}\cdot log(\hat{p}_v)+(1-y_{PL_v})\cdot log(1-\hat{p}_v)].%L_{BCE}(p_v,PL_v) 
\end{equation}
% where the label selection mask $m_v$ is the same definition as in~\cite{DPL}.
% \begin{equation}
% m = 1[]
% \end{equation}

\begin{table*}[htb]
\caption{Quantitative comprehensive comparisons of performance on \textbf{adversarial and clean} images on the test data averaged over all domain and over all OC and OD segmentation. C+O denotes overall results averaged on the compound and open target domains.}\label{tab2}
\centering
\scalebox{0.73}{
\begin{tabular}{l|c|c|c|c|c|c|c|c|c|c}
\hline \multirow{2}{*}{Method} & \multicolumn{4}{|c|}{ Compound(C) } & \multicolumn{2}{c|}{ Open(O) } & \multicolumn{4}{c}{ Avg. } \\
\cline { 2 - 11 } & \multicolumn{2}{|c|}{ RIM-ONE-r3 } & \multicolumn{2}{c|}{ Drishti-GS } & \multicolumn{2}{c|}{ REFUGE val } & \multicolumn{2}{c|}{ C } & \multicolumn{2}{c}{$\mathrm{C}+\mathrm{O}$} \\
\cline { 2 - 11 } & $\mathrm{D}[\%]$ & ASD & $\mathrm{D}[\%]$ & ASD & D[\%] & ASD & $\mathrm{D}[\%]$ & ASD & $\mathrm{D}[\%]$ & $\mathrm{ASD}$ \\

\hline
% W/o adaptation & \textcolor{green}{\ding{52}} &	7.4700	&–	&–&	80.22&	13.4709&	–&	–&	–&	–\\

% Oracle\cite{BEAL} & \textcolor{red}{\ding{56}} &	5.8935&	–&	–&	90.10&	11.4964&	–&	–\\

AdvEnt \cite{ADVENT}	& 	69.64&  	25.58&82.71&  	13.86&	46.60 & 	50.11&	76.17 & 	19.72&	66.31 & 	29.85
\\

BEAL \cite{BEAL} &	62.56&	22.46 & 78.00 &	16.00 &	41.67&	50.44&	70.28&	15.47&	60.74&	27.13
\\
OCDA\cite{open}&	69.55&32.66 &83.64&13.41 &	62.67&41.62 &	76.59&23.03 	&71.95&29.23 
 \\
DPL \cite{DPL} &		73.00&	23.68&87.23&	10.35&	62.92&	34.33&	80.11&	17.02&	74.38&	22.79
 \\

TT-SFUDA \cite{TTSFUDA} &		67.02&	37.45&	82.89	&13.82&63.22&	38.45&	74.95&	25.64&	71.04&	29.91
 \\

TENT\cite{TENT}  &		65.82&	35.26&84.42&	13.29 &	67.62&	30.58&	75.12	&24.27&	72.62 &	26.38
\\
SFDA-FSM~\cite{SFDA-FSM} &   	71.10&	26.61 	&84.87&	13.25&69.03&	23.11 	&77.99&	19.93 &	75.00&	20.99 
\\
% \begin{tabular}[l]{@{}l@{}}Ours\\ (Standard)\end{tabular} 
Ours (Standard Source) & 75.13&24.49 &	86.42&11.25 &76.57&17.36 &	80.78&17.87 &	79.37&17.70 
 \\

Ours (Robust Source) &\underline{76.14}&\underline{23.24} &	\underline{88.04}&\underline{9.22} 	&	\textbf{83.92}&\underline{12.01} &	\underline{82.09}&\underline{16.23} &	\underline{82.70}&\underline{14.82} 

\\

\textbf{Ours (Both)}  &	\textbf{76.79}&\textbf{16.59} 	& \textbf{88.85}&\textbf{8.52} &\underline{83.81}&\textbf{10.66} &	\textbf{82.82}&\textbf{12.55} &	\textbf{83.15}&\textbf{11.92} 
 \\

\hline
\end{tabular}}
\end{table*}

\subsubsection{Pseudo-Boundary Loss} 
In semantic segmentation, accurate boundary prediction between objects is critical. To encourage the model to predict more accurately, we use pseudo-boundaries obtained from a source model during the training. Explicitly modeling the boundaries helps the model generalize better to unseen target domains and generate more precise and coherent predictions, which can further improve the accuracy. Additionally, boundaries of foreground objects are domain-invariant information in different domains.

The pseudo boundary segmentation loss is defined as:
\begin{equation}
L_{bl} = \frac{1}{N}\sum_v(y_{PB_v}-\hat{b}_v)^2.
\end{equation}
Note that using robust pseudo labels and pseudo boundaries from a robust model can result in less accuracy, as the clean accuracy of the standard model is generally higher than that of the robust model. 
We wish to transfer the source robustness using a robust source model, on the other hand, we require better pseudo-labels to generate adversarial examples. Therefore, to balance the trade-off between clean and robust accuracy, we use both robust and standard source models and transfer them to the target domain. In Ours \textit{(Both)} method we obtain the required non-robust pseudo labels and pseudo boundaries using the standard source model $f_s$, then adapt the robust source model $f_s^r$ to the target domain. We use the non-robust parts from $f_s$ and prediction from $f_t^r$ to calculate the loss, as shown in in Figure \ref{fig-twostages}.(b). And only using standard source model $f_s$ (Ours (\textit{Standard Source})) is shown as Standard Flow in Figure ~\ref{fig1}, while only using robust source model $f_s^r$ (Ours (\textit{Robust Source})) is not shown but the only difference with standard flow is that only robust source model is available.

\subsubsection{Entropy Minimization} 
Entropy minimization is a useful technique for UDA~\cite{SRDA,TENT}, and minimizing entropy increases the confidence of the network's output. We apply this technique in our method to further improve the adaptation performance and encourage the target model to output confident predictions on unlabeled data. The Shannon entropy for a prediction probability is defined as:
\noindent 
\begin{equation}
L_{ent} = -\sum_v \hat{p}_v \cdot \log \hat{p}_v.
\end{equation}

$ \hat{p}_v$ denotes the mask probability obtained from the target model. To optimize the target segmentation model $f_t$, the overall loss function is:
\noindent 
\begin{equation}
L = L_{seg} +\alpha L_{bl} +\beta L_{ent},
\end{equation}
where $\alpha$ and $\beta$ are hyper-parameters for balancing the effect between the pseudo-boundary loss and entropy minimization loss. %Target training is described in Figure ~\ref{fig1}. 
% The details of training process are described in Fig.~\ref{fig1}.

%------------------------------------------------------------------------
\section{Experiments}
\subsection{Datasets}
We utilize four public optic disc (OD) and optic cup (OC) segmentation datasets. Specifically, we use the training set of the REFUGE challenge~\cite{Refuge} as the source domain, RIM-ONE-r3~\cite{RIM} and Drishti-GS~\cite{Drishti} as the compound target domains, and the testing set of REFUGE val~\cite{Refuge} as the open domain. The source domain consists of 400 annotated training images, and two compound target domain data are split to 99/60 and 50/51 images for training/testing respectively, following the same setup in DPL~\cite{DPL} and BEAL~\cite{BEAL}. The open domain consists of 80 images.

\begin{figure*}[htbp]
\centering\includegraphics[width=0.96\linewidth]{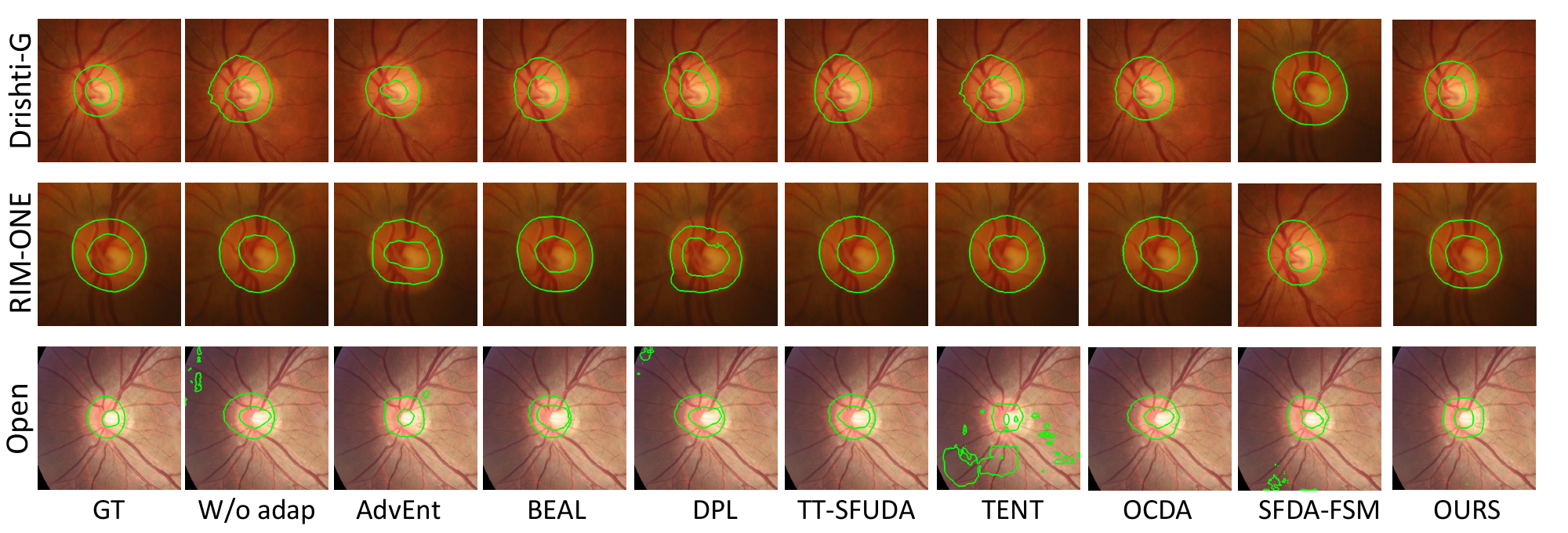}
   \vspace{-0.2cm}
\caption{Comparison of adaptation performance of different methods on \textbf{clean} samples.} \label{fig3}
\end{figure*}

\vspace{-0.25cm}
\begin{figure*}[htbp]
  \centering
  \vspace{-0.25cm}
   \includegraphics[width=0.96\linewidth]{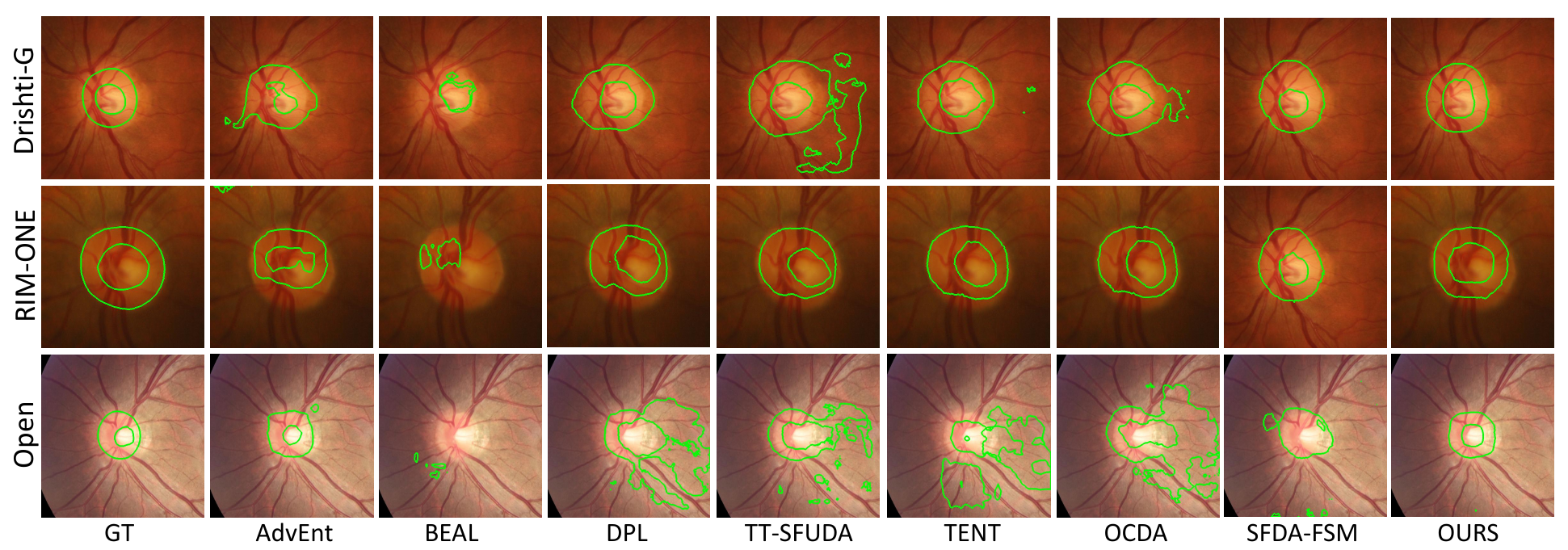}
   \vspace{-0.2cm}
   \caption{Comparison of adaptation performance of different methods on \textbf{adversarial} samples.}
   \label{fig3-adv}
\end{figure*}
%-------------------------------------------------------------------------
\subsection{Performance Metrics}
We employ two metrics to evaluate the segmentation performance: Dice coefficient and Average Surface Distance (ASD). 
% For clean and robust samples segmentation evaluation, w
The Dice coefficient measures pixel-wise segmentation accuracy, while ASD measures segmentation boundary accuracy. 
% Similarly, for robust segmentation evaluation, we use robust Dice and robust ASD, which is tested on adversarial samples. 
Higher Dice coefficients and lower ASD indicate better segmentation performance. 

%-------------------------------------------------------------------------
\subsection{Implementation Details}\label{details}
We use standard data augmentation, including Gaussian noise, contrast adjustment, and random erasing, same with~\cite{DPL} to slightly disturb the inputs when training target models, in order to make the predictions deviate from pseudo labels. And we also use these augmentation strategies in the standard source model training stage. For all methods, we use a MobileNetV2 adapted DeepLabv3+~\cite{DEEPLAB} as the backbone, similar to the work in~\cite{DPL}. The threshold $t$ is 0.75, same with BEAL~\cite{BEAL}. We train the source and target model both with batch size 8, but 200 epochs and 2 epochs respectively. The framework is implemented with Pytorch 1.12.1 using one NVIDIA GeForce RTX 2080 Ti GPU.
For generating adversarial images, we set the number of PGD steps to 20 and $\epsilon$ to 4/255. The loss components weights are set as $\alpha =1.0$, $\beta =0.4$.

%-------------------------------------------------------------------------

\subsection{Comparisons with the State-of-the-Art}
We conducted extensive experiments to compare our method with recent UDA methods, including BEAL~\cite{BEAL}, DPL~\cite{DPL}, AdvEnt~\cite{ADVENT}, TT-SFUDA~\cite{TTSFUDA}, TENT~\cite{TENT} and OCDA~\cite{open} on fundus image datasets. 
BEAL~\cite{BEAL} is an UDA method with adversarial learning between source and target data with boundary prediction on cross-domain fundus image segmentation; DPL~\cite{DPL} is an state-of-the-art SFUDA method with denoised pseudo labeling, which is the best-reported SFUDA model on cross-domain fundus image segmentation; AdvEnt~\cite{ADVENT} is a popular UDA benchmark approach that encourages entropy consistency between the source and target domains; TT-SFUDA~\cite{TTSFUDA} is a state-of-the-art two-stage approach for source-free domain adaptive image segmentation, including target-specific adaptation and task-specific adaptation; TENT~\cite{TENT} is a popular test-time adaptation method which optimizes the model by minimizing entropy of its prediction; OCDA~\cite{open} is a memory-based curriculum learning framework that improves generalization on the compound and open domains.1) a curriculum domain adaptation strategy to bootstrap generalization across domain distinction in a data-driven self-organizing fashion and 2) a memory module to increase the model's agility towards novel domains. 
SFDA-FSM\cite{SFDA-FSM}is a two-stage state-of-the-art two-stage approach for source-free domain adaptive image segmentation including a Fourier Style Mining (FSM) generation stage and an adaptation stage which has a Contrastive Domain Distillation (CDD) module to achieve feature-level adaptation.
%Since we follow the same network backbone and data split as BEAL and DPL, we reproduced the performance reported in their paper for comparison.
The detailed comparison results are presented in Table~\ref{tab1}, where we also include “\textbf{W/o adaptation}” lower bound and the supervised upper bound “\textbf{Oracle}” in the target domain. %(see Fig.~\ref{fig2}).
% Different domain adaptation methods can generally improve the performance over baseline, with 
% SFUDA method DPL~\cite{DPL} performs the best in fundus image segmentation.
While SOTA DPL performs well on clean samples, our two best approaches (Robust source and Both) significantly improve adversarial performance while maintaining clean performance. Notably, without source data during adaptation, our method (Both) achieves the highest Dice for optic disc segmentation on three target domains.

% As shown in Table 7 in supplementary, our method (Both) outperforms the baseline model with $82.82\%$ in C Dice, $12.55$ in C ASD and $83.15\%$ in C+O Dice, $11.92$ in ASD, improving C Dice by $12.54\%$ and C ASD by $2.92$, C+O Dice by $22.41\%$ and C+O ASD by $15.21$ compared to the baseline BEAL model.
The quantitative comparisons of different methods averaged on clean and adversarial images are shown in Table~\ref{tab2} and the Dice and ASD are averaged over OC and OD segmentation. Ours \textit{(Both)} outperforms all SF-DA methods on clean samples, especially the state-of-the-art source-free DPL by improving C+O Dice by $3.22\%$. % and C+O ASD by $1.98$. 
Our method (Both) also outperforms all UDA methods on adversarial samples, especially the baseline source-dependent BEAL by improving C Dice by $23.44\%$ and C+O Dice by $37.72\%$. % and C+O ASD by $16.64$. 
Moreover, when the gap between the source and target domain is trivial, as Drishti-GS, using non-robust (Both) or robust pseudo parts (Robust source) does not cause a big difference. Otherwise, non-robust counterparts are suggested.

Apart from closed-set UDA, we also demonstrate the effectiveness of our method for the open domain. As shown in Table \ref{tab2}, the proposed PLPB achieves the best accuracy and even outperforms the standard UDA method\cite{open}. The improvements for open-set tasks are significant, as low-level boundary information of the source model is utilized sufficiently to address the distribution shift and the robust model generalizes well on out-of-distribution data.

We again conduct experiments of all the baseline approaches with adversarial training\cite{PGD} in the source domain, as shown in Table S1 in Supplementary. Ours \textit{(Both)} still have the best overall performance compared with robust methods. And it is important to note that the clean metrics of other robust methods drop considerably if they are directly trained robustly, which is in line with the general observation that robust models tend to hurt the performance on clean samples\cite{robustness}.

Figure ~\ref{fig3} shows the qualitative comparison of adaptation performance on clean samples. Other methods hardly predict accurate boundaries in ambiguous regions and our method produces more accurate boundaries and mask predictions. Comparison of adversarial samples is shown in Figure \ref{fig3-adv}. Other methods have collapsed segmentation results on adversarial samples, while our PLPB remains well shapes thanks to robust augmentation and boundary information from the robust source model.

% Both method show better performance thanks to non-roubst pseudo labels and pseudo boundaries.
These results demonstrate the superiority of our method to adapt the model even without the source data.
% , due to the use of the denoised pseudo-labels, pseudo-boundary and entropy minimization to facilitate the learning from target samples.
This also indicates that SF-DA would not necessarily underperform the vanilla UDA. 
One possible reason is that Vanilla UDA methods assume that the source and target domains share common features, which is not always the case in real-world scenarios. Moreover, vanilla UDA methods aim to find an invariant latent space between the source and target distribution which could be challenging. Our method directly adapts the model to the target domain, allowing it to capture more discriminative representations from target distribution and be more robust to domain shifts. 

%-------------------------------------------------------------------------
\subsection{Ablation Studies}
We study the effectiveness of key components in PLPB, with results shown in
Table \ref{tab5} and Table \ref{tab6} for clean and adversarial samples respectively. Adding boundary loss significantly improves the source-only model. The entropy loss also contributes to improvements in terms of Dice score and ASD value. Also, the comparison with Ours (Standard) and Ours (Robust) in Table \ref{tab1} shows the superiority of using both standard and robust models. The results of the ablation studies confirm the effectiveness of the components in our method.

\begin{table}[ht]
\caption{Ablation results with different losses on \textbf{clean} samples. 
% $L1,L2, L3$ denotes $L_{seg}$, $L_{seg_b}$, $L_{ent}$ respectively
}\label{tab5}
\centering
\scalebox{0.58}{
\begin{tabular}{l|c|c|c|c|c|c|c|c|c|c}
\hline \multirow{2}{*}{ Method } & \multicolumn{4}{|c|}{ Compound(C) } & \multicolumn{2}{c|}{ Open(O) } & \multicolumn{4}{c}{ Avg. } \\
\cline { 2 - 11 } & \multicolumn{2}{|c|}{ RIM-ONE-r3 } & \multicolumn{2}{c|}{ Drishti-GS } & \multicolumn{2}{c|}{ REFUGE val } & \multicolumn{2}{c|}{ C } & \multicolumn{2}{c}{$\mathrm{C}+\mathrm{O}$} \\
\cline { 2 - 11 } & $\mathrm{D}[\%]$ & ASD & $\mathrm{D}[\%]$ & ASD & D[\%] & ASD & $\mathrm{D}[\%]$ & ASD & $\mathrm{D}[\%]$ & $\mathrm{ASD}$ \\

\hline
$L_{seg}$   &	83.85&9.88 &  89.30&8.12	&85.14&7.72& 	86.57&9.00 &	86.09&8.57 

 \\
$L_{seg}+ \alpha L_{bl}$  &84.12&9.27 & 89.76&7.79&	85.76&7.26 	&86.94&8.53 &	86.55&8.10 
\\
$ L_{seg}+ \beta L_{ent} $  &	84.08&	8.75&89.93&	7.84 &	85.30&	8.33 &	87.00&	8.30 &	86.44&	8.31 

\\

% $+$boundary loss & 89.76&7.79 &	81.92&10.73 &	85.76&7.26 &	85.84&9.26 	&85.81&8.59 $L_{seg}+\alpha L_{seg_b} +\beta L_{ent}$
% $L_{seg}+ \alpha L_{bl}+ \beta L_{ent}$ & \textbf{90.04}&\textbf{7.56}	&\textbf{85.42}	&\textbf{8.30}&\textbf{86.42}&\textbf{6.83}&	\textbf{87.73}&\textbf{7.93 }&	\textbf{87.29}&\textbf{7.56}

\begin{tabular}[l]{@{}l@{}}$L_{seg}+\alpha L_{bl} $\\ $+ \beta L_{ent}$\end{tabular}&\textbf{85.42}	&\textbf{8.30} & \textbf{90.04}&\textbf{7.56}	& \textbf{86.42}&\textbf{6.83}&	\textbf{87.73}&\textbf{7.93 }&	\textbf{87.29}&\textbf{7.56 }\\
\hline
\end{tabular}}
\end{table}

\begin{table}[ht]
\caption{Ablation results with different losses on \textbf{adversarial} samples. }\label{tab6}
\centering
\scalebox{0.58}{
\begin{tabular}{l|c|c|c|c|c|c|c|c|c|c}
\hline 
\multirow{2}{*}{ Method } & \multicolumn{4}{|c|}{ Compound(C) } & \multicolumn{2}{c|}{ Open(O) } & \multicolumn{4}{c}{ Avg. } \\
\cline { 2 - 11 } & \multicolumn{2}{|c|}{ RIM-ONE-r3 } & \multicolumn{2}{c|}{ Drishti-GS } & \multicolumn{2}{c|}{ REFUGE val } & \multicolumn{2}{c|}{ C } & \multicolumn{2}{c}{$\mathrm{C}+\mathrm{O}$} \\
\cline { 2 - 11 } & $\mathrm{D}[\%]$ & ASD & $\mathrm{D}[\%]$ & ASD & D[\%] & ASD & $\mathrm{D}[\%]$ & ASD & $\mathrm{D}[\%]$ & $\mathrm{ASD}$ \\

\hline
$L_{seg}$   &	67.98&29.84&  86.45&10.20 &	81.08&17.45  &	77.22&20.02 &	78.50&19.66 
 \\
$ L_{seg}+ \alpha L_{bl} $  &	68.03&26.60&86.98&9.97 &	81.09&18.33 &	77.50&18.29 &	78.70&18.30 

\\
$ L_{seg}+ \beta L_{ent} $ 	&68.57&	28.72 &86.38&	11.61 	&80.93&	18.86& 	77.47&	20.16 &	78.62&	19.73 

\\
% $L_{seg}+ L_{bl}+ L_{ent}$
\begin{tabular}[l]{@{}l@{}}$L_{seg}+ \alpha L_{bl} $\\ $+\beta L_{ent}$\end{tabular} &	\textbf{68.17}&\textbf{24.88} & \textbf{87.66}&\textbf{9.47} &	\textbf{81.20}&\textbf{14.49}& 	\textbf{77.91}&\textbf{17.17} &	\textbf{79.01}&\textbf{16.28}

\\
\hline
\end{tabular}}
\end{table}

%-------------------------------------------------------------------------

\subsection{Discussion}
Based on our experimental evaluations, we find that consistent with previous works\cite{advtl,r1}, robust source feature extractors are effective in preserving robustness, while maintaining sufficiently high accuracy on clean samples. And robust pre-trained models not only perform well on targets without adversarial training but also improve the accuracy on clean samples. 

Furthermore, adversarial training on source training enhances the ability of adaptation by improving the low-level diversity and generalization on the open dataset. We assume that adversarial attacks are based on the gradient of the training model, which can be considered as a type of low-level diversity so that adversarial samples can also serve as augmentation for diversity. It is consistent with the previous work that a pre-trained model that is more robust to input perturbation provides a better initialization for generalization on downstream out-of-distribution data~\cite{ood}.

After disabling the pseudo-boundary loss, the performance of the target model has dropped apparently. By explicitly modeling the boundaries in the target domain, our target model achieves better generalization and generates more precise and coherent predictions.

%------------------------------------------------------------------------
\section{Conclusion}
In this study, we develop a method for robust source-free domain adaptation for the segmentation of fundus images. Our method PLPB outperforms competing state-of-the-art methods, achieving SOTA performance on adversarial and clean samples. A limitation of our method is that we need to train two source models, which increases training costs. In the future, we will explore how to balance accuracy and robustness with a single source model. We will further explore the source domain distribution information embedded in the source model to jointly work with the pseudo-labeling to cope with a more severe domain shift. %\textcolor{red}{without altering source training.}

{\small
\bibliographystyle{ieee_fullname}
\bibliography{egbib}
}

\end{document}